\documentclass[pmlr]{jmlr}% new name PMLR (Proceedings of Machine Learning)

\usepackage{longtable}% for long tables
\usepackage{graphicx}
\usepackage{graphics}
\usepackage{arydshln}
\usepackage{comment}
 \usepackage{booktabs}
 \usepackage{multirow}
 \usepackage{etoolbox}

\usepackage[load-configurations=version-1]{siunitx} % newer version

\makeatletter
\def\set@curr@file#1{\def\@curr@file{#1}} %temp workaround for 2019 latex release
\makeatother

\theorembodyfont{\upshape}
\theoremheaderfont{\scshape}
\theorempostheader{}
\theoremsep{\newline}

\jmlrvolume{}
\jmlryear{}
\jmlrworkshop{}

\title{An Extensive Data Processing Pipeline for MIMIC-IV}

\author{
    \centerline{Mehak Gupta,
    Brennan Gallamoza,
    Nicolas Cutrona,
    Pranjal Dhakal,}\\
    \centerline{Raphael Poulain,
    Rahmatollah Beheshti}\\
    \Email{\centerline{\{mehakg, brennang, ncutrona, dpranjal, rpoulain, rbi\}@udel.edu}}\\
    \addr 
       \centerline{University of Delaware, Newark, Delaware, USA}
}

\begin{document}
\raggedbottom
\maketitle

\begin{abstract}
  An increasing amount of research is being devoted to applying machine learning methods to electronic health record (EHR) data for various clinical purposes. This growing area of research has exposed the challenges of the accessibility of EHRs. MIMIC is a popular, public, and free EHR dataset in a raw format that has been used in numerous studies. The absence of standardized pre-processing steps can be, however, a significant barrier to the wider adoption of this rare resource. Additionally, this absence can reduce the reproducibility of the developed tools and limit the ability to compare the results among similar studies. In this work, we provide a greatly customizable pipeline to extract, clean, and pre-process the data available in the fourth version of the MIMIC dataset (MIMIC-IV). The pipeline also presents an end-to-end wizard-like package supporting predictive model creations and evaluations. The pipeline covers a range of clinical prediction tasks which can be broadly classified into four categories - readmission, length of stay, mortality, and phenotype prediction. The tool is publicly available at \url{https://github.com/healthylaife/MIMIC-IV-Data-Pipeline}.
\end{abstract}

\section{Introduction}
\label{sec:intro}
The rapid increase of the adoption  of electronic health records (EHRs) for recording patients' data across different healthcare systems worldwide, along with the swift pace of applying machine learning (ML) methods to study various problems in this domain has led to a surge of ML tools that work on top of EHRs. Despite the great potential and huge interest, there remain noticeable barriers in front of both technical and biomedical communities to create impactful and reliable tools that can meaningfully improve individuals' health. Among the primary barriers is having access to large EHR datasets. Due to various concerns, such as interoperability, safety, and privacy; easy access to EHRs remains a major challenge for researchers and practitioners in the community. 

In this context, the MIMIC initiative has been functioning as a pioneer and prominent large-scale openly available EHR dataset that addresses the privacy issues of EHR data by carefully de-identifying patient information \citep{mimicold}. The latest (major) version of this dataset is version four (MIMIC-IV), which was offered in 2020 and iteratively updated since then \citep{mimic}. 
% Extracted from the Beth Israel Deaconess Medical Center in Boston, MA, USA,  MIMIC consists of over 257,000 patients' demographics, diagnoses, labs, medications, and vitals.

Despite being easily accessible, tasks surrounding data cleaning and pre-processing in MIMIC remain challenging, setting a relatively high bar to enter the field. This issue is particularly important as working with EHR datasets often requires interdisciplinary  technical and clinical knowledge. The MIMIC initiative states that ``data cleaning steps were not performed, to ensure the data reflects a real-world clinical dataset'', and therefore ``[r]esearchers should follow best practice guidelines when analyzing the data'' \citep{mimic}. Poor study designs, by not following the aforementioned ``best practice guidelines,''  can lead to unreliable, biased, or even harmful designs. A few examples of some noticeable problematic designs using MIMIC have been recently presented by \cite{boag2022ehr}. Also, the lack of a standardized pipeline to extract and pre-process MIMIC-IV limits the comparability and reproducibility of different research work \citep{johnson2017reproducibility,mcdermott2021reproducibility}. While several cohort extraction and pre-processing pipelines exist for MIMIC (reviewed in the next section), they mostly focus on standardizing specific and predetermined processes and are less concerned about presenting a flexible, user-defined pipeline that follows a vetted process. Accordingly, there is a need for a standardized data pre-processing pipeline that is simple to use, customizable according to user preferences, and flexible with respect to defining cohorts used in different tasks. 

% Another advantage of an open-source pre-processing pipeline for MIMIC-IV is that it reduces time needed to clean, pre-process and use MIMIC-IV. Any research study with experiments focused on proprietary datasets \citep{gupta2021concurrent,gupta2022flexible,gupta2022obesity} can easily incorporate MIMIC-IV  with our pipeline to create reproducible work.

In this paper, we present a pipeline targeted toward the above gap, offering a configurable framework to prepare MIMIC-IV data for downstream tasks. The pipeline cleans the raw data by removing outliers and allowing users to impute missing entries. It provides options for the clinical grouping of medical features using standard coding systems for dimensionality reduction.  It can produce a smooth time-series dataset by binning the sequential data into time intervals and allowing for filtering of the time-series length according to user preference. 
% For example, if a user wants to extract a cohort for the first 24 hours of all admissions with the hourly time-series resolution, data outside of the user-defined time boundaries will be excluded and 24 hours of data can be binned into equal time intervals of one hour each. 
All of these options are customizable, allowing users to generate a personalized  patient cohort. Importantly, the customization steps can be recorded and shared, increasing the reproducibility of the studies using this pipeline. Besides the main pre-processing parts, our pipeline also includes two additional parts for modeling and evaluation. For modeling, the pipeline includes several commonly used ML and deep learning sequential models for performing prediction tasks. The evaluation part offers a series of standard methods for evaluating the performance of the created models and includes options for evaluating the fairness and interpreting the models. Such a pipeline can increase the usability of MIMIC by making it more accessible to researchers, and can significantly reduce the time and experience needed to clean, pre-process and use the dataset. Considering the growing number of studies that use MIMIC-IV (around 300 in mid-2022), we believe an efficient pre-processing pipeline can allow many more research teams to seamlessly incorporate this dataset into their work. 

\section{Related Work}
\label{sec:relw}
Among the major openly available EHR datasets, including eICU \citep{pollard2018eicu}, HiRID \citep{hird,yeche2021hirid}, All of Us \citep{RN351}, and the synthetic repositories like Synthea \citep{synthea}, MIMIC is the most widely used EHR dataset containing ICU and non-ICU patient data. As MIMIC offers raw EHR data, the community has come up with several suggested procedures for working with the data. Among these efforts, a few aimed to standardize the cleaning, pre-processing, and representation of the earlier versions of MIMIC. Additionally, most of these efforts aimed at presenting a standardized benchmark for certain prediction tasks along with the description of the data processing pipeline. Similar to other standardized benchmarking efforts in the ML community, one primary goal of these studies was to allow follow-up studies to improve the presented baseline performances.  For MIMIC-III, these benchmarks have been proposed by \cite{harutyunyan2019multitask}, \cite{purushotham2018benchmarking}, \cite{sjoding2019democratizing}, \cite{wang2020mimic}, and \cite{tang2020democratizing}. The two benchmarks proposed by \cite{harutyunyan2019multitask}, and \cite{purushotham2018benchmarking} provide data pre-processing steps for mortality and length of stay (LOS) prediction tasks using 17 and 148 selected variables from the dataset, respectively. The more popular work by \cite{harutyunyan2019multitask} focuses  on data inclusion and exclusion criteria and is less devoted to pre-processing and cleaning steps for a larger set of features from MIMIC data. MIMIC-EXTRACT  is another popular pipeline for MIMIC-III that uses a larger set of variables and includes ventilation, vasopressors, and fluid bolus therapies,  expanding the tractability for downstream tasks \citep{wang2020mimic}. The focus of benchmark-oriented studies generally is not related to providing flexibility to users with respect to customizable feature selection  or user-defined pre-processed cohorts for desired downstream tasks. This can be seen in MIMIC-EXTRACT \citep{wang2020mimic}, as the pipeline only aggregates data into fixed hourly time windows; or in the work by \cite{harutyunyan2019multitask} and \cite{purushotham2018benchmarking} that only provide a few selected variables. 

Besides studies solely devoted to MIMIC, there is also a larger family that targets EHR data in general. A notable pipeline of this type is FIDDLE \citep{tang2020democratizing}, which presents a pre-processing pipeline to generalize over the MIMIC-III and eICU datasets. It includes pre-processing steps that remove redundant features based on the underlying distribution of the data. Though this step assists with reducing dimensionality, it does not consider clinical domain knowledge, which could potentially lead to  clinically relevant features being removed in practice. As another example for this family, the extensive framework presented by \cite{jarrett2020clairvoyance} provides a general pipeline that is not specific to a certain EHR dataset. The pipeline does not assist in cohort extraction steps but provides different pre-processing and imputation techniques for time-series medical data.

While closely relevant, existing pipelines for prior versions of MIMIC (such as MIMIC-III) cannot be directly used for MIMIC-IV for a multitude of reasons. One glaring difference is the use of ICD-9 diagnosis codes in MIMIC-III to define cohorts, versus using both ICD-9 and ICD-10 diagnosis codes in MIMIC-IV. 

A recently proposed pipeline for MIMIC-IV is COP-E-CAT \citep{mandyam2021cop} that can configure pre-processing steps, providing a user some flexibility to define cohorts as needed for analysis tasks. The main limitation of this work is that it only considers ICU data, leaving out non-ICU data. The end tasks defined mainly consist of mortality prediction, electrolyte repletion, and dimensionality reduction, leaving out  more popular tasks such as readmission, and LOS.
%  and phenotype prediction for different conditions. It provides dimensionality reduction using an algorithm but does not consider clinical knowledge for the task, which could lead to the loss of clinically relevant features. 

% In this work, we propose an end-to-end pipeline that works with the MIMIC-IV EHR dataset for both ICU and hospital data (non-ICU). We provide a very flexible end-to-end pipeline that can be used to define cohorts for readmission and mortality prediction by including all admission records. Additionally, we provide users the possibility to define cohorts for - readmission, LOS, mortality, and phenotype prediction  which can be further refined to the four most prevalent chronic conditions, where admission records are filtered according to the specific disease-related diagnosis at admission time. Our pipeline significantly improves existing pipelines by providing automated clinical groupings for diagnoses and medication codes. 
% Furthermore, we provide an exhaustive list of evaluation metrics for EHR prediction tasks, including fairness evaluation metrics with respect to patient demographics.

\begin{figure*}[t]
  \centering 
  \includegraphics[width=\textwidth,height=4in]{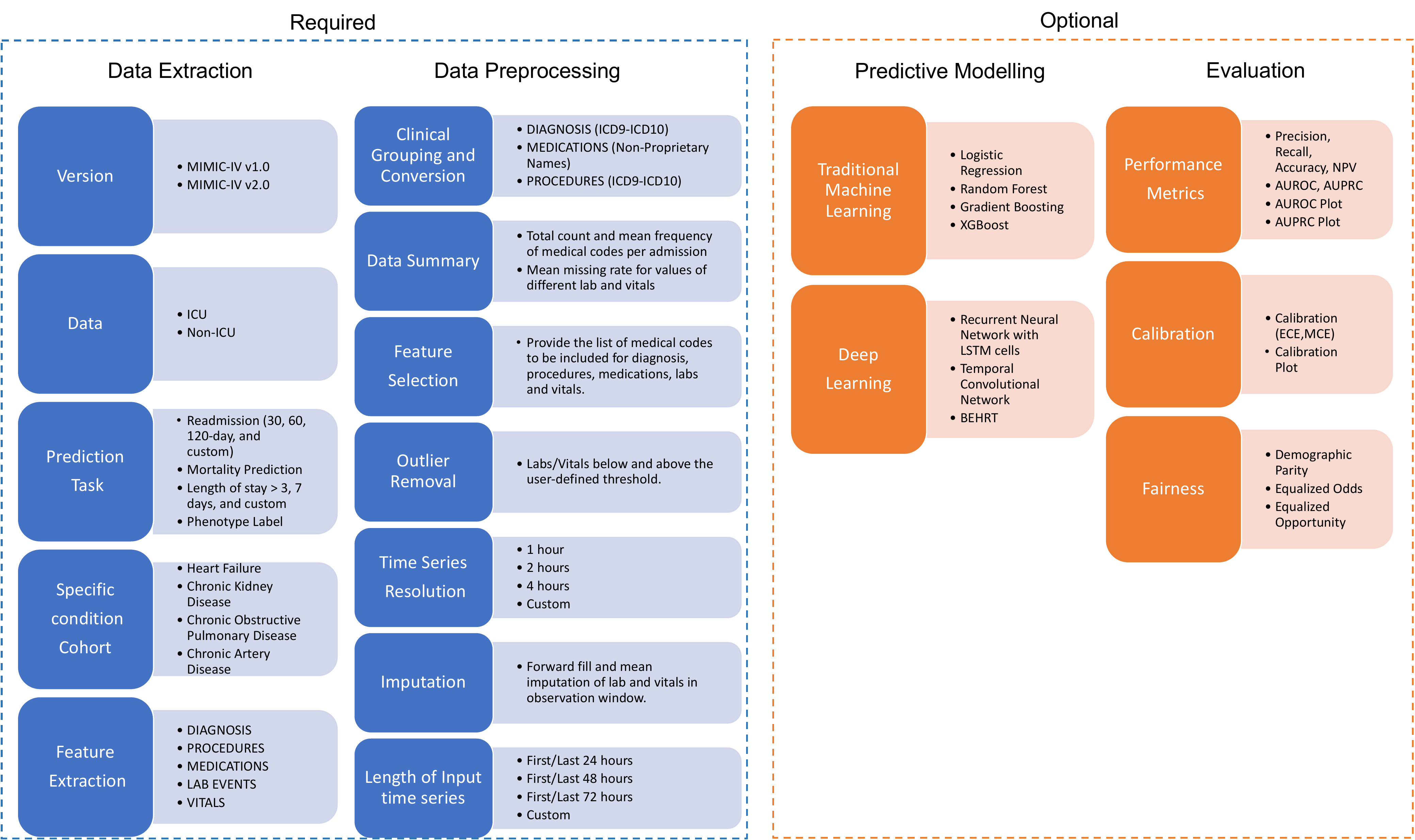} 
  \caption{Pipeline overview. The left two parts are required to produce the processed data. }
  % Data Extraction and Preprocessing parts are required to produce the output (processed) data. Data Modelling and Evaluation parts are optional and can be used if needed for training and evaluating the models provided in the pipeline. Evaluation can also be used as a standalone part to obtain performance metric scores for results obtained using any data and machine learning model.}
  \label{fig:example} 
\end{figure*} 

\section{MIMIC-IV Dataset}
MIMIC (Medical Information Mart for Intensive Care)-IV comprises de-identified EHRs from the patients admitted at Beth Israel Deaconess Medical Center in Boston, MA, USA, from the years 2008 to 2019. The data contains  information for each patient regarding hospital stay, including laboratory measurements, medications administered, and, vital signs. It consists of data for over 257,000 distinct patients, yielding 524,000 admission records.
MIMIC-IV is separated into \lq\lq modules\rq\rq\  to reflect the provenance of the data. We use three modules in our pipeline: core, ICU, and hosp, consisting of patient stay information, ICU-level data, and hospital-level data respectively. Our pipeline is compatible with both v1.0 and v2.0 (the latest) of MIMIC-IV. MIMIC-IV v2.0 only contains  ICU and hosp modules, and consists of all the tables found in v1.0.

\section{Pipeline Description}

Our pipeline contains four main parts: data extraction, data pre-processing, predictive modeling, and model evaluation as shown in Figure \ref{fig:example}. The pipeline aims to offer a flexible package that identifies commonly used/recommended practices and also supports user-defined choices for preparing the raw data. 

\subsection{Data Extraction}
% The goal of the data extraction part is to select a cohort of patients for a prediction task following a user-specified set of criteria. 
The data extraction part allows the extraction of cohorts for both ICU and non-ICU data. A user can choose a specific prediction task from the list of in-hospital mortality, readmission, LOS, or phenotype prediction using ICU and non-ICU data. All prediction tasks can be further refined for four major chronic health conditions: heart failure, chronic kidney disease (CKD), chronic obstructive pulmonary disease (COPD), and coronary artery disease (CAD) to create 16 different prediction tasks. To extract a cohort for a certain condition, we include admission records that have diagnoses related to the chosen chronic condition at the time of admission. To identify which admissions have been diagnosed with specific conditions, we use the International Classification of Diseases Version 10 (ICD-10) code categories. Details about ICD-10 selection criteria are  discussed in Appendix \ref{apx:icd10}. Using ICD-10 codes enables non-medical expert users (including ML practitioners without access to subject matter experts) to use the openly available ICD-10 directory to identify higher-level codes for specific conditions \textsc{(disease\_cohort.py)}. 

Readmission tasks can be defined for readmission after user-defined (between 10 to 150) days from the previous admission discharge time. LOS prediction tasks are defined as LOS greater than user-defined (between 1 to 10) days. Unlike the retrospective phenotype labeling defined by \cite{harutyunyan2019multitask}, we define phenotype label prediction task to predict labels for four major chronic conditions for the next visit. MIMIC only records diagnosis at the start of every hospital admission, therefore phenotype label prediction can be performed for the next admission based on the current admission. Readmission and phenotype prediction tasks are defined using the last 24, 48, 72, and custom hours of admission data. Similarly, the mortality and LOS prediction task can be defined for prediction after the first 24, 48, or 72, custom hours and 12, 24, custom hours of the admission respectively \textsc{(day\_intervals\_cohort.py)}.

\begin{figure}[t]
  \centering 
  \includegraphics[width=\columnwidth]{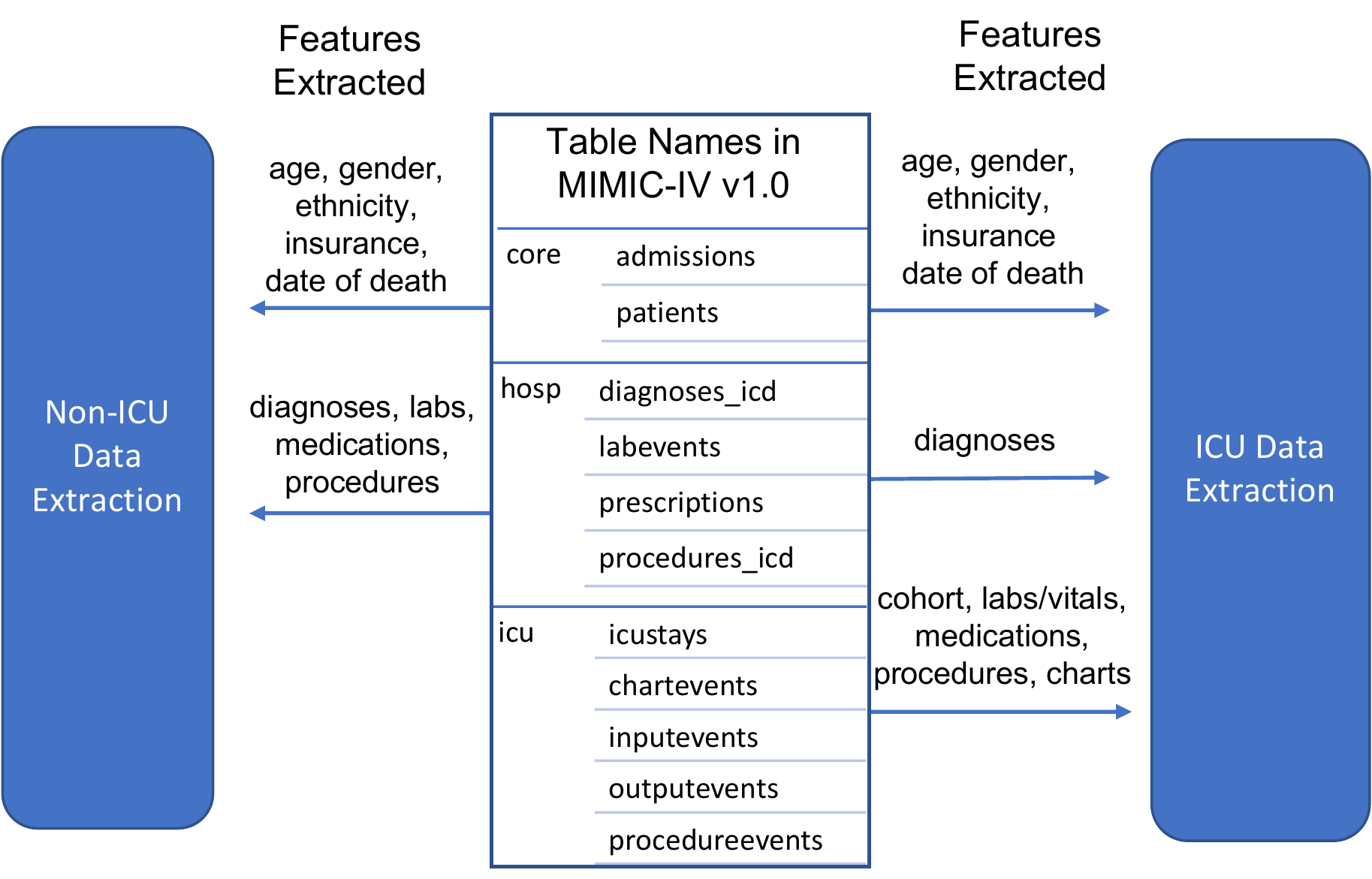} 
  \caption{Example extracted features.}
  \label{fig:data} 
\end{figure}

As part of the data extraction, the pipeline also extracts the features for each cohort. One can select from diagnoses, procedures, medications, labs, and vitals among the available clinical data associated with each admission. In addition to the time-series data such as labs, vitals, and medications, the pipeline also includes demographic data that contains information regarding age, gender, insurance, and ethnicity for each admission record. As shown in Figure \ref{fig:data}, a user can extract different features from the available tables in MIMIC-IV. Due to the fundamental differences in pediatric problems, for all cohort extractions, the pipeline only includes adult patients (18 years or older). To calculate the age in readmission and mortality prediction, we subtract a patient’s admission time from their year of birth. A patient’s year of birth is obtained by subtracting their “anchor\_year” from their “anchor\_age”. 
% Using this calculation, we added age information to our cohort and also ensured that a patient must be 18 years old by the start of their recorded admission time.

\subsection{Data Pre-processing}
The data pre-processing part consists of the steps to configure the representation of EHR data for the downstream tasks. It consists of (1) clinical grouping, (2) feature summarizing and selection, (3) outlier removal, and (4) time-series representation  \textsc{(feature\_selection\_icu.py} , \textsc{feature\_selection\_hosp.py)}.

\paragraph{Clinical Grouping}
Clinical grouping helps reduce data dimensionality while preserving important clinical information. The pipeline provides clinical grouping with respect to diagnosis and medications. Prior to any grouping of diagnosis data, the ICD-9 codes are converted to ICD-10, so that all codes can be grouped by the same standard. The pipeline follows the method presented by \cite{rasmy2020representation} for mapping ICD-9 to ICD-10 codes and uses the first three digits as the root of ICD-10 codes to group the diagnosis codes. 

The clinical grouping of medications is performed by converting drug names into their non-proprietary counterparts through the National Drug Codes (NDCs) index \citep{NDC}. As a first pre-processing step, NDCs in the MIMIC-IV dataset and the FDA’s public NDC database were both converted into 11-digit forms. With the formatted NDC codes, the MIMIC-IV medication data were mapped to the FDA’s NDC database of non-proprietary drug names through the label and product portions of the NDC codes (digits 1–5 and 6–9, respectively).

\paragraph{Feature Summarizing and Selection} After grouping, the pipeline provides summaries of the features extracted. These summaries are provided in CSV files, which contain information about the mean frequency of each medical code per admission and also the percentage of missing values for different labs and medications. The user can refer to these summaries and decide which medical codes they want to include for each feature in this step. As an example, for labs, the pipeline provides a summary containing the mean frequency of all the different lab codes and the percentage of missing values associated with each lab code. Based on this information, the user can then provide a list of lab codes to keep in the data in a CSV file. If the user opts to keep all codes, they do not need to specify any list.

\paragraph{Outlier removal}
Unlike the outlier removal in COP-E-CAT \citep{mandyam2021cop}, where only values labeled 999999 and the negative (infeasible) values are removed, the proposed pipeline identifies statistically high or low values for each feature distribution separately. The pipeline uses the user-defined threshold percentile values as boundary points for each feature and either removes or replaces the values outside of that interval. For example, if the user chooses a threshold of 2, the pipeline either removes the values above the 98th and below the 2nd percentile or replaces them with the 98th and 2nd percentile values. Before performing outlier detection, the pipeline also performs unit conversion to make sure all the values for a feature are measured using the same unit of measurement \textsc{(outlier\_removal.py)}.

\begin{figure}[t]
  \centering 
  \includegraphics[width=\columnwidth]{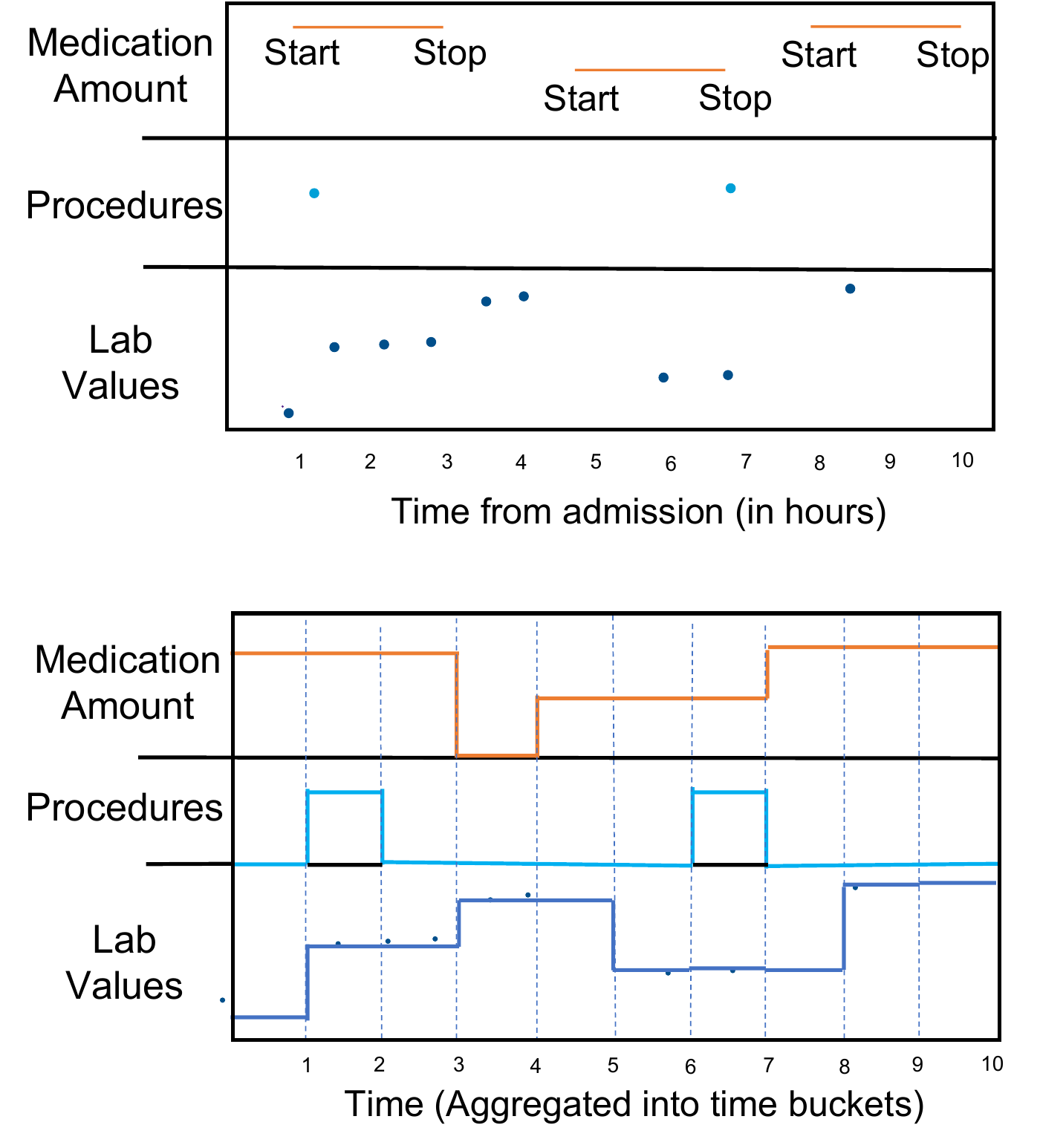} 
  \caption{Processing Raw EHR. }
  \label{fig:smooth3} 
\end{figure}

\paragraph{Time-series Representation} In this step, the pipeline lets the user first define the observation window which is the duration of time for which input features will be observed before making predictions. The pipeline then bins the data into uniform time intervals for each time-series feature. These time intervals can be specified by the user, representing the time-series resolution. For every binned time interval, the pipeline labels a feature as 1 if it is recorded within that time interval and 0 otherwise.  By default, the values for labs and vitals are forward-filled to adjacent time intervals.  If the values are missing for a specific feature, the pipeline uses the mean value of the feature for data imputation. Here, the imputation is performed after data is selected for the observation window. Thus, our imputation technique only uses values in the observation window to impute other values and prevents any data leakage from the values outside of the observation window. The user can choose not to perform imputation using forward filling and mean imputation. In the case of no imputation, unmeasured values in the time intervals will be represented by a 0. No imputation is done for medications; instead, every time interval between the start and stop times is denoted by their respective dosage. Time intervals where medications are not administered are labeled with a  0. This process is shown in Figure \ref{fig:smooth3}. It also shows that after pre-processing the data for labs/vitals, procedures, and medications, the data is converted into a continuous time series. Missing labs/vitals are forward filled and high-frequency values are aggregated in different regularized time intervals. Similarly, medication data also represents the complete time signal by their dosage between the start and stop time if a medication was administered and 0 otherwise. Procedures are represented by 1 in the time interval where the procedure is performed and 0 otherwise. Since diagnoses are recorded at the time of admission and do not change over the course of a patient's hospital or an ICU stay, they are represented as a static feature for each admission \textsc{(data\_generation.py, data\_generation\_icu.py)}.

Throughout the pre-processing steps, the pipeline performs various data verification checks to prevent any misinterpretations during model learning. It excludes admission records where admit time is greater than discharge time. It also removes any medications with a start time that takes place after a stop time, as well as medications that have a start time after the patient's discharge time. To keep medication times consistent, it shifts the start time of medication to admit time if the medication was given  before admission and continues into a patient's stay. It also shifts the stop time to discharge time if the medication stop time is after the patient's discharge time.

\paragraph{Output Format} After completing data pre-processing and representing the data as time-series signals, the pipeline saves the data in three CSV files for each sample. Each admission record will have three sets of CSV files for static features, dynamic features, and demographic features. The data saved as a CSV is in a 2-dimensional format of (\#time, \#features). A dictionary output is also saved, which has all of the feature data recorded for a cohort in a single pickled file. A detailed example of a dictionary structure is shown on our GitHub page \textsc{(data\_generation.py, data\_generation\_icu.py)}.

\subsection{Predictive Modelling}
The third (optional) part of the pipeline is to construct a model to train for a  prediction task. Users can also build their own models and use the data saved in the previous steps of the pipeline for training. Our data in CSV files can be easily used to format data for non-time series and time series models. The pipeline includes several common architectures: logistic regression, random forest, gradient boosting, XGBoost,  recurrent neural network with LSTM cells \citep{hochreiter1997long}, temporal convolutional neural network (TCN) with 1D (one-dimensional) fully conventional network \citep{lea2017temporal}, and a common transformer model used for EHRs called BEHRT \citep{li2020behrt}.  For the non-neural network models, a user has the option to either concatenate the values of the features over time to form a 2D output (\#samples, \#time-points, \#features), or aggregate feature values over time to form a 2D output (\#samples, \#features). For the neural network models, time-series data is formatted into three dimensions as (\#samples, \#time, \#features) and static data in 2-dimensional format as (\#samples, \#features). In basic versions of LSTM and TCN architectures, static data is repeated for each timestamp. In the hybrid models, a separate feed-forward network is used for static data. For BEHRT, we take extra pre-processing steps to convert the time-series sequence of events to a sequence of tokens as expected by the existing implementation of BEHRT. More details about the model architectures are provided in Appendix \ref{apx:model}. During the predictive modeling step, the pipeline also provides the option for k-fold cross-validation and randomly oversampling the minority classes \textsc{(ml\_models.py, dl\_train.py)}.

\subsection{Model Evaluation}
The evaluation part of the pipeline, consists of an extensive list of performance metrics, allowing for easy study comparison. This optional part provides scores for commonly used metrics such as AUROC, AUPRC, accuracy, precision, recall, and NPV. It also includes calibration metrics using Expected Calibration Error (ECE) and Maximum Calibration Error (MCE). Calibration metrics help evaluate the degree to which a model is certain about the predicted outcome \citep{matheny2019artificial} and quantify how well the predicted probabilities of an outcome match the probabilities observed across the data  \textsc{(evaluation.py)}. 

The pipeline also includes a dedicated fairness auditing component, targeting age, gender, and ethnicity as the sensitive attributes. To measure a model's fairness, quantitative metrics including demographic parity, equalized opportunity, and equalized odds are used   \textsc{(fairness.py)}

\begin{table}[t]

\centering 
  \caption{Experimental results for six example tasks of readmission, mortality, LOS$>$3 days, and phenotype prediction for heart failure for patients admitted with chronic kidney disease, using  ICU and non-ICU data (positive to negative ratio).} 
  % The cohorts for readmission and phenotype tasks have been defined to include ``Last 48 hours" of data, the mortality tasks include ``First 48 hours" of data, and LOS$>$3 includes ``First 24 hours" of data. The time resolution is 2 hours for all cohorts.}
  \label{tab:icu}
  \resizebox{\columnwidth}{!}{
  \centering
  \begin{tabular}{l|cc|cc|cc|cc|cc|cc|cc|cc}
    \toprule
    Task$\rightarrow$&\multicolumn{4}{c|}{Readmission}&\multicolumn{4}{c|}{Mortality}&\multicolumn{4}{c|}{LOS$>$3 days}&\multicolumn{4}{c}{Heart Failure in next 30 days}\\ 
     \toprule
    Data$\rightarrow$&\multicolumn{2}{c|}{ICU (0.19)}&\multicolumn{2}{c|}{Non-ICU (0.27)}&\multicolumn{2}{c|}{ICU (0.12)}&\multicolumn{2}{c|}{Non-ICU (0.04)}&\multicolumn{2}{c|}{ICU (0.45)}&\multicolumn{2}{c|}{Non-ICU (0.55)}&\multicolumn{2}{c|}{ICU (0.18)}&\multicolumn{2}{c}{Non-ICU (0.24)}\\ 
     \toprule
    Metric$\rightarrow$&AUROC&AUPRC&AUROC&AUPRC&AUROC&AUPRC&AUROC&AUPRC&AUROC&AUPRC&AUROC&AUPRC&AUROC&AUPRC&AUROC&AUPRC\\
    \midrule
    Logistic Regression&0.57&0.23&0.58&0.34&0.67&0.24&0.73&0.15&0.64&0.59&0.66&0.70&0.53&0.20&0.55&0.28\\
     Random Forest&0.60&0.25&0.68&0.56&0.79&0.39&0.80&0.45&0.72&0.70&0.72&0.72&0.55&0.22&0.60&0.29\\
 
     Gradient Boosting&0.62&0.27&0.72&0.62&0.85&0.48&0.84&0.52&0.74&0.74&0.80&0.78&0.58&0.23&0.64&0.33\\
     XGBoost&0.60&0.25&0.74&0.63&0.84&0.47&0.85&0.53&0.75&0.74&0.83&0.81&0.62&0.23&0.65&0.33\\
     
     LSTM (Time-Series)  &0.72&0.45&0.77&0.64&0.86&0.49&0.85&0.60&0.74&0.73&0.83&0.82&0.75&0.45&0.70&0.54\\
     TCN (Time-Series)&0.71&0.43&0.75&0.63&0.84&0.46&0.84&0.58&0.72&0.71&0.79&0.79&0.73&0.41&0.68&0.52\\
 
     LSTM (Hybrid)&0.74&0.45&0.79&0.65&0.87&0.49&0.86&0.60&0.76&0.75&0.85&0.82&0.76&0.46&0.72&0.56\\
     TCN (Hybrid)&0.72&0.43&0.74&0.63&0.85&0.47&0.84&0.59&0.74&0.71&0.80&0.79&0.75&0.41&0.70&0.53\\

  \bottomrule
\end{tabular}
}
\end{table} 

\subsection{Using the pipeline}
The pipeline is available in the form of sequential wizard-style code blocks with a simple (Jupyter) Notebook interface to interact with the user. To begin, the user should start the \textsc{mainPipeline.ipynb} Notebook, which guides the user through each sequential step. It also provides instructions to use the evaluation part as a standalone part. As an open-source tool, the user can use the pipeline and choose from the available options or add new pieces at any stage.

\section{Proof of concept experiments}\label{sec:exp}
We conducted a series of experiments to demonstrate how the pipeline can be used. We extracted a cohort from MIMIC-IV v1.0, corresponding to the patients who had chronic kidney disease at the time of admission. For feature selection, we use only diagnosis and labs/vitals as input features and further select the features for labs/vitals from the list provided by \cite{wang2020mimic}.

We set the time-series data to include the ``Last 48 hours" of data for the 30-day readmission task and phenotype prediction, ``First 48 hours" of data for the mortality prediction, and ``First 24 hours" of data for the LOS$>$3 days prediction in both ICU and Non-ICU data. The data resolution for time-series data is set to 2 hours. 

Additional experimental settings, including the training setup and recorded pipeline choices, are presented in Appendix \ref{apx:experiments}. The hyperparameters for deep learning models are set using the \textsc{(parameters.py)} file. Table \ref{tab:icu} shows the obtained AUROC and AUPRC in these experiments. We also evaluated the fairness performance of one of the models (the LSTM model for the mortality prediction task), as shown in Table \ref{tab:viz_report} in Appendix \ref{apx:fairness}.

\section{Discussion} 

% Through this work, we provide a pipeline to clean and pre-process the MIMIC-IV EHR dataset for use in analysis tasks with machine learning models. In Section \ref{sec:exp} via four prediction tasks, we demonstrate how our proposed pipeline can be used to extract and pre-process MIMIC-IV EHR data to model and evaluate deep learning methods built using the selected data. We show how different preferences can be selected to define a customized cohort while recording the choices made for future reproducibility.

An increasing amount of research is being performed in the area of machine learning to build clinical predictive models on EHR data. The MIMIC dataset promotes research in this area by  making a large EHR dataset available to researchers. In this paper, we provided an open-source standardized pipeline to clean and pre-process MIMIC-IV data for use in ML models. The MIMIC initiative serves as a major step toward democratizing access to EHRs, and we consider the presented pipeline as a step toward democratizing access to MIMIC. The ease of use of this pipeline makes MIMIC-IV data more accessible to both clinical and non-clinical researchers working on EHR prediction modeling. 

Existing pipeline tools only extract and pre-process a limited number of features for a specific downstream prediction task, providing limited options for users to define a customized cohort as discussed in Section \ref{sec:relw}. While these benchmark-centered tools address the issue of reproducibility and comparability, they do not assist researchers in creating different studies with different cohort requirements. Our proposed pipeline is highly configurable and provides users with many options to define customized cohorts by allowing for feature selection options and other user-defined pre-processing steps. Our pipeline not only addresses the issue of reproducibility (by recording all design choices) but also promotes further research using different cohorts for prediction tasks.

% The optional standalone evaluation part provided in our pipeline extends the usability of our work beyond the MIMIC-IV dataset. This part can be used to evaluate the results of any dataset and any machine learning model, given the user can provide model output (logits or predicted probabilities) and ground truth labels.

\paragraph{Limitations:}

Our study trades off a more generalizable pipeline applicable to different versions of MIMIC or different EHR datasets for a more specialized version focusing on MIMIC-IV, with the goal of lowering the barriers to entering the field. While our pipeline is fully compatible with the latest MIMIC version, we expect that small adjustments will make that compatible with future versions as well. Legacy versions of MIMIC (such as MIMIC-III) continue to be largely used by the community and this would likely be the case for MIMIC-IV, too. Due to a predefined set of prediction tasks, our pipeline could be less relevant for tasks significantly different from the included tasks.  In the future, we aim to extend our pipeline by explicitly adding more prediction and regression tasks, such as predicting the time for the next admission, and future diagnosis codes. We also plan to extend our pipeline evaluation part by adding evaluation for regression and multi-label prediction tasks. Moreover, by publicly offering our pipeline, we hope that others can also join us in the continuous improvement of this pipeline. 

% ACKNOWLEDGEMENTS ONLY GO IN THE CAMERA-READY, NOT THE SUBMISSION
% \acks{Many thanks to all collaborators and funders!}

\bibliography{main}

\appendix
\section{ICD-10 codes to identify admissions with specific conditions}\label{apx:icd10}
ICD-10 codes are 3-7 character identifiers that specify
diagnoses, where the first 3 characters represent a code’s category, giving a broader definition
of the diagnosis. We use the first 3 characters in the ICD-10 diagnosis recorded at the admit time to label admissions based on heart failure, CKD, COPD, or CAD. With respect to heart failure, we linked this condition with the category I50, which maps to heart failure (WHO), and labeled this admission with I50 at admit time. Similarly, for CKD, COPD, and CAD we linked N18, J44, and I25 respectively in the first three characters of ICD-10 diagnosis at admitting time.

\section{Predictive Model Description}\label{apx:model}
\begin{figure*}[htb]
  \centering 
  \includegraphics[width=\textwidth]{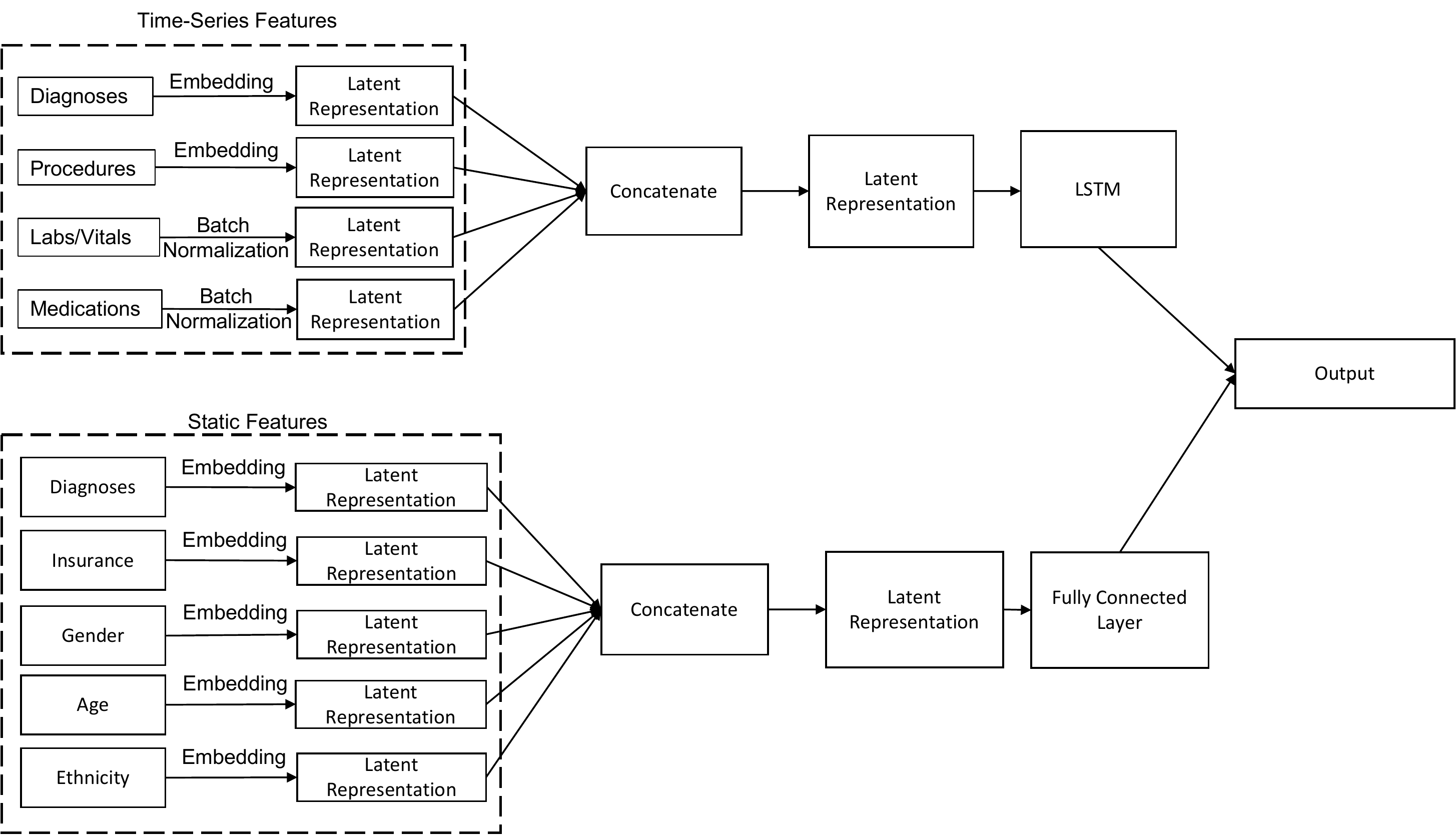} 
  \caption{Model Architecture}
  \label{fig:model} 
\end{figure*}

LSTM models are designed using two stacked unidirectional LSTM layers with a dropout of 0.2. TCN models use convolution layers with a kernel size of 10 and a stride of 1 along with batch normalization. Figure \ref{fig:model} shows the overall architecture of our models for both LSTM and TCN. It shows that all time-series features go through their own separate LSTM or TCN layers. All static features go through their own separate fully-connected layer network. The reason for designing models with separated networks for each feature is that it provides flexibility to add or remove features by switching on or off the specified network. Each variable goes through an embedding layer before being fed to its respective network. Time-series models only use time-series features and hybrid models use both time-series and static features. Output from all separate LSTM/TCN/fully connected layers are concatenated at the end to collectively go through another fully connected network to produce the output label. For implementing the BEHRT, the pipeline creates a separate code block to pre-process the data files for use in existing BEHRT implementation. We bin the numerical data into pre-defined quantiles as expected by BEHRT implementation. The maximum number of events per sample is recommended at 512 and therefore, we remove samples that have more than 512 event sequences to effectively use the existing implementation.

\section{Experiment setup}\label{apx:experiments}

We perform 5-fold cross-validation and use 80\% of the data for training and 20\% for testing. Another 10\% of the samples in the training data are randomly extracted as the validation set. The final model is chosen based on the validation performance, which is then used to report performance on the test data. The experiments run for a maximum of 50 epochs until the validation loss stops improving for 5 epochs continuously. We use binary cross-entropy loss for training the models. 

Table \ref{tab:exp}  shows how one can record the choices made through the data extraction and pre-processing parts. It shows the summary of steps and choices made for the mortality prediction experimental setup. 

\begin{table}[!htbp]

\centering 
  \caption{Recording pipeline choices for reproducibility purposes. An example of defining data for the mortality prediction task.}
  \label{tab:exp}
  \resizebox{\linewidth}{!}{%
  \begin{tabular}{l|l}
    \toprule
    Data Extraction&Data Pre-processing\\
    \midrule
  1. MIMIC-IV v1.0&1. Group Diagnoses codes for with ICD-10\\
2. ICU&2. Produce data summary for features\\
    3. Mortality&3. Feature selection for Labs/Vitals\\
        4. Admitted for Chronic Kidney Disease&4. Outlier removal with threshold .98\\
        5. Diagnosis and Labs/Vitals&5. 2 hour time-series resolution\\
        &6. Forward fill and mean imputation for labs/vitals\\
        &7. First 48 hours of admission data\\
    
  \bottomrule
\end{tabular}
}
\end{table} 

\section{Fairness Evaluations}\label{apx:fairness}
Table \ref{tab:viz_report} shows the output of the fairness evaluation for LSTM model for non-ICU mortality prediction task. This report shows how the output of the fairness part is displayed in the pipeline. Thie fairness report is saved in a CSV file for later use. . For demographic parity, equalized opportunity, and equalized odds to hold, the Positive Rate (PR), True Positive Rate (TPR), and True Positive and False Positive Rate (TPR and FPR) need to be equivalent across different sub-groups within each sensitive attribute. Specifically, for demographic parity to hold, the Positive Rate (PR) should be equivalent across different sub-groups within each sensitive attribute. We see that the PR for sub-groups  \textit{White} (0.0102) in ethnicity is higher compared to PR for \textit{African American} (0.0024). Therefore, the model is more likely to predict a patient's mortality for the \textit{White} sub-group than the \textit{African American} sub-group, potentially leading to unfair treatment outcomes including the prioritization of resources for patients that are more likely to survive.  We can see from the Table \ref{tab:viz_report} that the TPR for \textit{male} (0.0574) and \textit{female} (0.0796) are similar in value. Thus, the model satisfies the equalized opportunity definition for gender. We can also observe that the FPR for \textit{male} (0.0057) and \textit{female} (0.0064) are similar in value. Because of this similarity, the model satisfies the equalized odds notion for gender because both TPR and FPR values are comparable. Equalized odds implies equalized opportunity, enforcing equivalent FPR values across different sub-groups. 

\begin{table}[t]
    \centering
    \caption{Fairness Evaluation Report}
  \label{tab:viz_report}
  \resizebox{\columnwidth}{!}{
 \begin{tabular}{c|c|c|c|c|c|c}
\toprule
\textbf{Sensitive} &Groups&  \textbf{TPR} &    \textbf{TNR} &    \textbf{FPR} &    \textbf{FNR} &     \textbf{PR}  \\
\textbf{Attributes} &&&&&& \\
\toprule
          ethnicity&AMERICAN INDIAN &       NaN & 1.0000 & 0.0000 &    NaN & 0.0000  \\
                              &ASIAN &    0.0000 & 0.9949 & 0.0051 & 1.0000 & 0.0048  \\
             & AFRICAN AMERICAN &   0.0000 & 0.9976 & 0.0024 & 1.0000 & 0.0024 \\
                      &HISPANIC/LATINO &  0.1429 & 0.9973 & 0.0027 & 0.8571 & 0.0077  \\
                                &OTHER &   0.0909 & 1.0000 & 0.0000 & 0.9091 & 0.0033 \\
                  &UNABLE TO OBTAIN &    1.0000 & 1.0000 & 0.0000 & 0.0000 & 0.0588 \\
                              &UNKNOWN &   0.1481 & 0.9650 & 0.0350 & 0.8519 & 0.0529 \\
                                &WHITE &  0.0723 & 0.9928 & 0.0072 & 0.9277 & 0.0102 \\
          \toprule
                        Gender&          F &    0.0574 & 0.9943 & 0.0057 & 0.9426 & 0.0076 \\
                                   & M &  0.0796 & 0.9936 & 0.0064 & 0.9204 & 0.0100 \\
         \toprule
                           Age&  20-30 &    0.0000 & 0.9783 & 0.0217 & 1.0000 & 0.0215 \\
                            & 30-40 &    0.0000 & 1.0000 & 0.0000 & 1.0000 & 0.0000 \\
                              &40-50 &    0.0000 & 0.9981 & 0.0019 & 1.0000 & 0.0019 \\
                                &50-60 &   0.0526 & 0.9949 & 0.0051 & 0.9474 & 0.0065 \\
                             & 60-70 &    0.0986 & 0.9925 & 0.0075 & 0.9014 & 0.0112 \\
                                 &70-80 &   0.0667 & 0.9946 & 0.0054 & 0.9333 & 0.0086 \\
                               & 80-90 &   0.0947 & 0.9922 & 0.0078 & 0.9053 & 0.0128 \\
                               & 90-100 &  0.0000 & 0.9951 & 0.0049 & 1.0000 & 0.0047 \\
\bottomrule
\end{tabular}
}
\end{table}

\end{document}